\documentclass{article}

\usepackage{arxiv}

\usepackage{lineno}

\usepackage[utf8]{inputenc} 
\usepackage[T1]{fontenc}    
\usepackage{hyperref}       
\usepackage{url}            
\usepackage{booktabs}       
\usepackage{amsfonts}       
\usepackage{nicefrac}       
\usepackage{microtype}      
\usepackage{lipsum}
\usepackage{graphicx}
\usepackage{authblk}

\usepackage{amsmath}
\usepackage{multirow}
\usepackage[ruled,vlined,linesnumbered]{algorithm2e}

\title{QROSS: QUBO Relaxation Parameter optimisation via Learning Solver Surrogates}

\author[1]{Tian Huang}
\author[2]{Siong Thye Goh}
\author[2]{Sabrish Gopalakrishnan}
\author[1]{Tao Luo}
\author[1]{Qianxiao Li}
\author[2]{Hoong Chuin Lau}
\affil[1]{Institute of High Performance Computing, Agency for Science Technology and Research, Singapore}
\affil[2]{School of Computing and Information Systems, Singapore Management University, Singapore}

\begin{document}

\maketitle

\begin{abstract}
An increasingly popular method for solving a constrained combinatorial optimisation problem is to first convert it into a quadratic unconstrained binary optimisation (QUBO) problem, and solve it using a standard QUBO solver. However, this relaxation introduces hyper-parameters that balance the objective and penalty terms for the constraints, and their chosen values significantly impact performance. Hence, tuning these parameters is an important problem. Existing generic hyper-parameter tuning methods require multiple expensive calls to a QUBO solver, making them impractical for performance critical applications when repeated solutions of similar combinatorial optimisation problems are required. In this paper, we propose the QROSS method, in which we build surrogate models of QUBO solvers via learning from solver data on a collection of instances of a problem. In this way, we are able capture the common structure of the instances and their interactions with the solver, and produce good choices of penalty parameters with fewer number of calls to the QUBO solver. We take the Traveling Salesman Problem (TSP) as a case study, where we demonstrate that our method can find better solutions with fewer calls to QUBO solver compared with conventional hyper-parameter tuning techniques. Moreover, with simple adaptation methods, QROSS is shown to generalise well to out-of-distribution datasets and different types of QUBO solvers.
\end{abstract}

\section{Introduction} \label{sec:intro}

The Quadratic Unconstrained Binary Optimisation problem (QUBO) has become a unifying formulation for a wide range of combinatorial optimisation problems \cite{kochenberger2014unconstrained,anthony2017quadratic}. Its objective form is closely related to the Hamiltonian of the Ising model in statistical physics \cite{glover2018tutorial}, where a variety of sampling methods based on Markov Chain Monte-Carlo have been developed to probe the energy landscape and find its ground states \cite{brooks2011handbook}. These form the basis of commercially available QUBO solvers, such as those implemented on quantum annealers \cite{mcgeoch2014adiabatic} and quantum-inspired computers \cite{aramon2019physics}.

Many combinatorial optimization problems can be expressed in the form of $\min_{x\in \{0,1\}^n} x^TQx$ subject to $Cx=d$ where decision variables are binary and the constraints are linear. The conversion of such problem into a QUBO form is straightforward where the problem can be rewritten as $\min_{x\in \{0,1\}^n} x^TQx+A\|Cx-d\|^2$ \cite{glover2018tutorial}, where $A$ is referred as the relaxation parameter. 
The challenge is that the quality of the solutions found by a QUBO solver is sensitive to the choice of the parameter value. Inappropriate choice of relaxation parameter value could lead to a solution that is either infeasible or far from optimal. Hence, relaxation parameter optimisation is an important step in this problem-solving process.

In the context of emerging computing technology and new QUBO solvers, existing methods do not capture the common structure shared by instances of a problem. For industry applications like vehicle route planning \cite{goddard2017will} and resource allocation \cite{sao2019application}, instances of the same problem are solved repeatedly. We argue that there is valuable information in the history of solved instances that can be extracted as prior knowledge for a relaxation parameter optimisation method.

In this paper, we propose a QUBO Relaxation parameter Optimisation method based on QUBO Solver Surrogates (QROSS) to mitigate these issues. As indicated by its name, QROSS relies on a surrogate model that approximates important characteristics of a QUBO solver. We build the surrogate model using machine learning on data obtained from the QUBO solver when applied to solve instances of a problem. QROSS then proposes promising relaxation parameters using the surrogate model on new instances, effectively reducing the number of calls to QUBO solver.

The solver surrogate in QROSS only models certain aspects of the original QUBO solver that are necessary for relaxation parameter tuning. 
Given a problem instance and a relaxation parameter, the solver surrogate predicts the probability that the QUBO solver finds a feasible solution, as well as the objective energies. Note that no explicit solutions are predicted by the solver surrogate. An evaluation on the solver surrogate, which we will model as a carefully designed neural network, is much cheaper/faster than a call to a QUBO solver.

Based on the solver surrogate, we propose two search strategies that will provide the trial values of the relaxation parameters. Hence, given a new instance of a problem, QROSS can propose promising relaxation parameters that can be used to form the QUBO problem that is then solved by a QUBO solver.

QROSS has the following features that make it unique among other hyper-parameter tuning methods:
\begin{itemize}
    \item QROSS captures the common structure of a class of problems. This is achieved by learning a solver surrogate from a history of problems. Knowledge in previous problems help to solve new problems of the same type. With QROSS, users are expected to find better relaxation parameters and solutions with fewer calls to the QUBO solver;
    \item Given a new problem of the same class, QROSS is able to predict the landscape of the objective function and help users understand the expectations without resorting to the expensive QUBO solving step;
    \item QROSS allows trade-off between optimality and number of calls to a QUBO solver. If an application only allows one call per problem, QROSS produces one parameter candidate, which has a good chance to satisfy the constraints of the problem. If a user can afford more calls for a problem, QROSS can change the strategy and propose more candidates to find better parameters and solutions. 
\end{itemize}

Our contributions are as follows:

\begin{enumerate}
    \item We introduce QROSS, a machine learning method to extract knowledge from QUBO problem instances solved in the past to facilitate the relaxation parameter optimisation process for a new instance of the same problem.
    \item We propose two offline and one online parameter selection strategies. Offline strategies learn from instances in the past and propose promising parameters without calling QUBO solvers. The online strategy learn from the instance-to-solve and improves parameter by exploiting results from QUBO solvers.
    \item We take Traveling Salesman Problem (TSP) as a case study to demonstrate that QROSS learns knowledge from history effectively and proposes promising parameters.
\end{enumerate}

In experiments, we use Fujitsu \emph{Digital Annealer} (DA) as a QUBO solver and compare QROSS with representative optimisation methods. Results show that QROSS outperforms these methods by a large margin. We train QROSS on synthetic dataset and find it is also adequate on out-of-distribution and real-world problem settings. We repeat all experiments with another QUBO solver, the hybrid solver \emph{Qbsolv} from DWave Quantum Annealer. Results suggest that QROSS generalises well for the different QUBO solvers we used. 


\section{Related Work} \label{sec:related}

Many parameter-sampling methods have been proposed in the literature, for example, SMAC \cite{hutter2011sequential} uses random forest; GPyOpt \cite{gonzalez2016gpyopt} and Spearmint \cite{snoek2012practical} use Gaussian Process, and Hyperopt \cite{bergstra2013hyperopt} uses tree-structured Parzen estimator. Optuna \cite{akiba2019optuna} allows the user to dynamically construct the search space. 

While the above papers are generic in that they do not take problem specific features into consideration, SATzilla \cite{xu2008satzilla} is an automated approach for constructing per-instance algorithm portfolios for the Satisfiability Problem (SAT) that uses the so-called empirical hardness models to choose among their constituent solvers.
Instance-Specific Algorithmic Configuration (ISAC) \cite{malitsky2014instance} predicts the impact of different parameters and the performance by learning from instances. In \cite{ansotegui2016maxsat}, ISAC was proposed to choose appropriate MAXSAT problem. In \cite{cenamor2016ibacop}, a per-instance configurable portfolio which is able to adapt itself to every planning task was proposed.  

Parameter tuning aside, generic constraint programming solvers have been proposed, for example \cite{hentenryck2009constraint} and \cite{benoist2011localsolver}, that perform local search by taking the constraints explicitly into the consideration. 

All methods mentioned above are typically not concerned with feasibility of solutions, since feasibility is guaranteed through the algorithm itself. However feasibility is not guaranteed by a QUBO solver. Tuning QUBO relaxation parameters will need to take feasibility into consideration. 



\section{The QROSS Method} \label{sec:method}

QROSS exploits knowledge learned from past instances of a problem when solving a new instance. In this section, explain what is the knowledge, why is it useful, how to learn it, and finally how to exploit it.

\subsection{Motivation}

As noted in the Introduction, the choice of the relaxation parameter value directly impacts feasibility. In other words, the concern is whether a solution for the QUBO model instantiated with a particular parameter value can be found that satisfies the constraints of the original constrained problem. We first investigate the relation between feasibility and parameter value. We denote the relaxation parameter as $A$, a problem instance as $g$, and define the probability of feasibility as the proportion of feasible solutions produced by the QUBO solver. Eq.\ref{equ:pf} gives an estimate of it when evaluated on a large number of solutions on a common parameter setting: 

\begin{equation}
P_f \approx \frac{\textrm{number of feasible solutions}}{\textrm{number of solutions}}
\label{equ:pf}
\end{equation}

\begin{figure}[htb!]
    \centering
    \includegraphics[width=0.95\columnwidth]{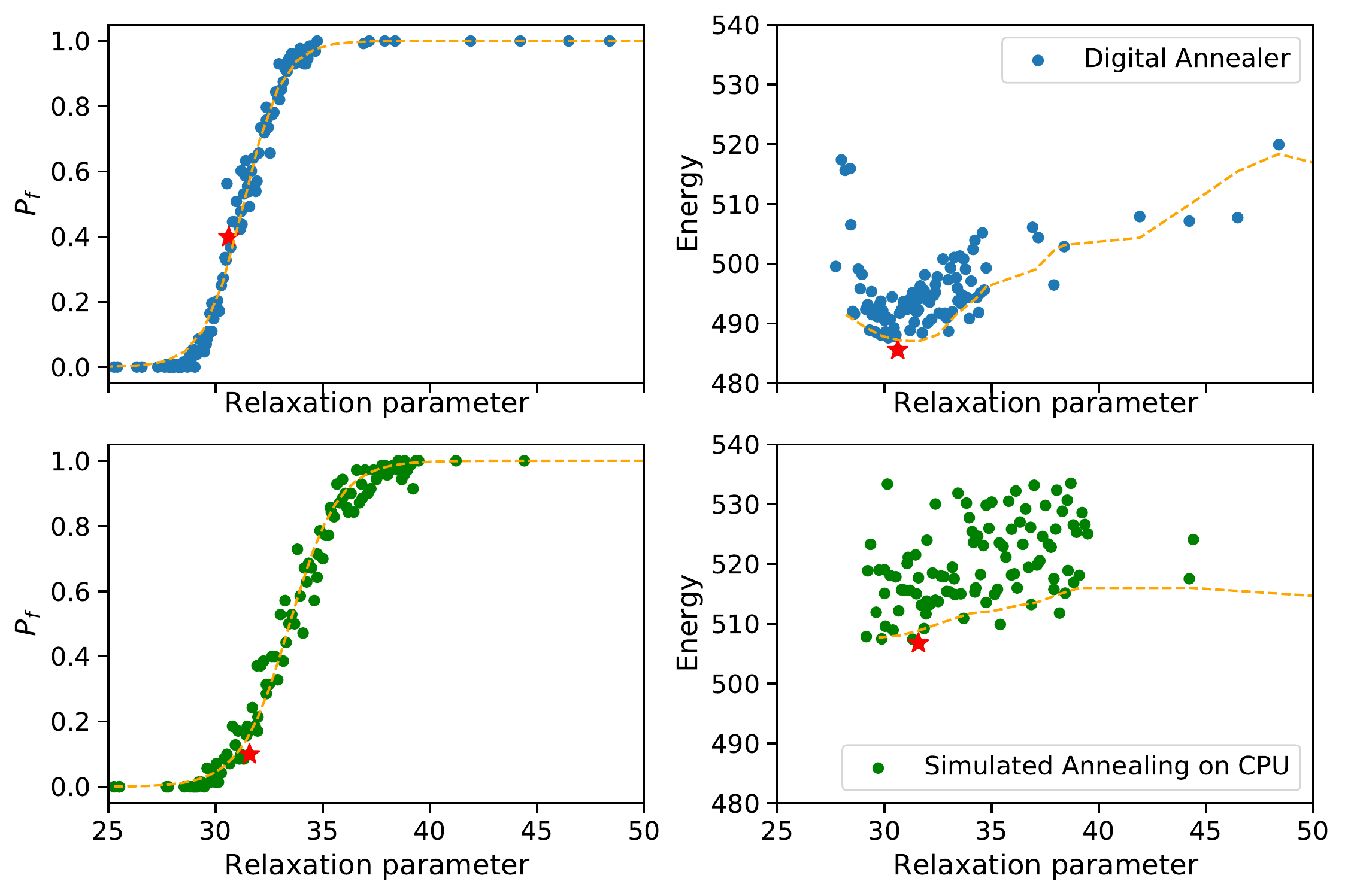}
    \caption{Probability of Feasibility and Objective Energy}
    \label{fig:pf_dmin}
\end{figure}

The value of $P_f$ lies in the range $[0, 1]$. The relation between $A$ and $P_f$ is shown in the left column of Fig.\ref{fig:pf_dmin}. The upper part of the figure shows the results from a QUBO solver, which is the Digital Annealer from Fujitsu \cite{aramon2019physics}. For each point, there are 128 solutions returned by the QUBO solver, given a problem instance $g$ and a relaxation parameter $A$. Because of the stochastic nature of QUBO solvers, the result could be different even if the input is the same. The dashed line that we fitted to the points looks like a sigmoid shape. The QUBO solver for the lower part of the figure is Simulated Annealing on CPU. The same shape can be observed from the lower part.

The relaxation parameter $A$ is also closely related to the objective energy of a solution. The upper right part of Fig.\ref{fig:pf_dmin} is produced by Fujitsu DA. Each point represents the minimum of fitness of 128 solutions from the QUBO solver, given an instance $g$ and a parameter $A$. The orange dashed line sketches the envelope of the points and presents as a dipper shape. The red star-shaped marker is the bottom of the dip, where the optimal solution appears. The parameter that leads to near optimal solution is called optimal parameter. The lower right part of the figure is for Simulated Annealing, the dipper shape is not as obvious as the upper part. This is because SA gets stuck in local minima more often for the current parameter setting and has an almost flat landscape.

We explain the intuition behind the observation as follows. As one increases the relaxation parameter value, the constraints, in the form of penalty, gradually dominates the QUBO objective. Thus, A QUBO solver will be more likely to find feasible solutions. Meanwhile, the part of the QUBO objective corresponding to the original objective becomes less prominent and sometimes even disappears. The QUBO solver will not be able to respond to the tiny difference presented by the objective part and fails to find (near-)optimal solutions. \footnote{This holds true for classical computers and quantum computers. Please refer to Appendix for detailed explanation.} On the other hand, if one decreases the relaxation parameter over a certain threshold, i.e., to the left part of the dip, there will be fewer feasible solutions, and therefore fewer chances for the QUBO solver to find a solution with a smaller fitness value. An analytic approximation of "expectation of minimum fitness" will be given in section \nameref{sec:MFS}, show that a promising relaxation parameter balances the weight of objective and feasibility.

Through these observations we arrive at the following \emph{Hypothesis}: Optimal solutions appear within $0<P_f<1$, i.e., on the slope of the Sigmoid shape. With more experiments we confirm this hypothesis holds true for all instances in TSPLIB \cite{reinelt1991tsplib} with Fujitsu DA and QAPLIB \cite{burkard1997qaplib} with SA on CPU. \footnote{We do not examine problems with size over 100 because solving these problems on hardware with limited memory requires decomposition into smaller problems, which is out of the scope of this paper.} Hence, if one can effectively predict $P_f$ and objective energy, the parameter optimisation process would be much easier. This is especially useful when the objective landscape is flat, because conventional parameter optimisation methods cannot effectively find clues from a flat landscape.

Fortunately, instances of a problems share common structure. For example, a car company has to do vehicle routing in a city many times a day. A logistic company has to manage \cite{goddard2017will} allocations in a warehouse repeatedly \cite{sao2019application}. For instances of a problem, the relation between relaxation parameter $A$, $P_f$ and objective energy are learnable. With these knowledge, we can select promising parameter values before calling the QUBO solver. The selection of parameter value does not involve exact-solution-finding process, thus it is much cheaper and faster than using the QUBO solver.

\subsection{Network Architecture}

Next we describe the network architecture for the surrogate. We have two inputs for the solver surrogate: the problem instance, denoted as $g$, and relaxation parameter, denoted as $A$. $A$ could be a scalar of vector, depending on the number of relaxation parameter of a problem. Learning from $g$ is tricky because the size of an instance could vary largely. We use an feature extraction layer that handle problems of different sizes. After the feature extraction, instances of different sizes are converted into fixed-size feature vectors. The feature vectors, together with the relaxation parameter $A$, are passed to a fully connected (FC) layer for further calculation.

We use the surrogate to predict two types of outputs. The first one is the probability of feasibility $P_f$, which is a function of $g$ and $A$. $P_f(g, A)$ is in the range $[0, 1]$. We can use sigmoid function as the activation of the last layer of the FC network and use Binary Cross Entropy loss (BCE) to learn $P_f(g, A)$. The other target is to predict fitness, i.e., the objective energy of a solution that satisfies the constraints of the original problem. Predicting the fitness is tricky. When there is no feasible solutions, there will be no fitness available for the instance, and this is problematic for neural network training. Our workaround is to predict the statistics of QUBO objective energy, $E_{avg}(g, A)$ and $E_{std}(g, A)$. $E_{avg}(g, A)$ and $E_{std}(g, A)$ are functions of $g$ and $A$. Any reasonable loss criteria for regression should work for the learning these functions. We use Huber loss as we are expecting many outliers in the dataset, due to the stochastic nature of a QUBO solver. Once we have $P_f(g, A)$, $E_{avg}(g, A)$ and $E_{std}(g, A)$, we can calculate the fitness using the analytical expression given in section \nameref{sec:MFS}. A detailed description of the architecture can be found in Appendix.

\begin{figure}[htb!]
    \centering
    \includegraphics[width=0.8\columnwidth]{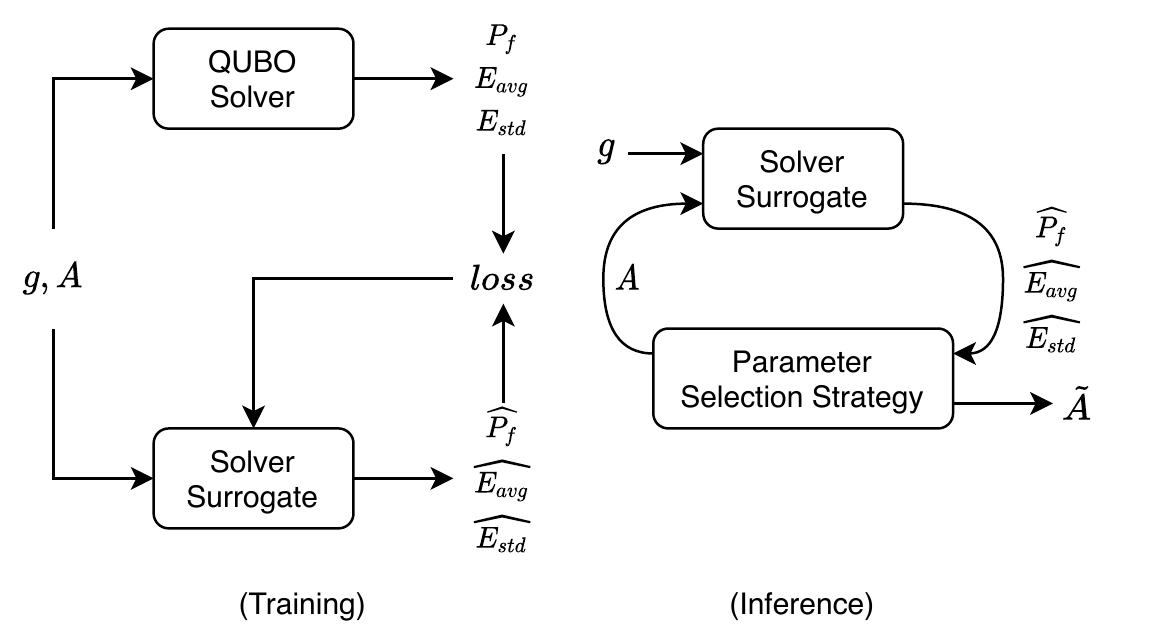}
    \caption{Training and Inference}
    \label{fig:train_test}
\end{figure}

The training workflow of the QROSS is described in the upper part of Fig.\ref{fig:train_test}. Given an instance $g$ and a parameter $A$, we find the prediction and the ground truth of $P_f$, $E_{avg}$ and $E_{std}$ by QUBO solver and solver surrogate. The difference between the prediction and the ground truth, i.e., the loss, can be used to train solver surrogate.

\subsection{Data Preparation} \label{sec:data_prep}

To train the solver surrogate, we prepare a dataset from past instances and solutions of a solver. Because heuristic QUBO solvers have a stochastic nature, it usually returns a batch of solutions and corresponding objective energy for an instance $g$ and parameter $A$. One can check the feasibility of each solution and calculate $P_f$ for the batch. We observed that the objective energy in a batch usually follows a bell-shaped distribution. Therefore, we use Gaussian sufficient statistics $E_{avg}(g, A)$, $E_{std}(g, A)$, i.e., mean and standard deviation, to characterise the objective energy.

A good coverage of sampling in terms of $A$ helps to improve training. Since the slope of the sigmoid shape is the region that we are interested in, we make sure that $\{A \; | \; 0<P_f(g, A)<1\}$ are well sampled for each problem. Including the plateau region of the sigmoid shape helps to prevent model overfitting and improves accruacy. Hence, we make sure that at least a sizable number of samples in $\{A \; | \; P_f(g, A)=0 \; \mathrm{or} \; 1 \}$ for each problem are included in the dataset. 

Data augmentation also helps the training. For example, pre-processing techniques, e.g. shifting or scaling, moves $A$ of different problems to the same order of magnitude so that learning and prediction become easier. Normalisation helps the convergence of the training curve. These techniques are applied in our experiments. 

\subsection{Inference} \label{sec:inference}

Next, we describe how to exploit the knowledge in the surrogate. The diagram of the inference is shown in the lower part of Fig.\ref{fig:train_test}. Given an instance $g$ and a parameter $A$, the solver surrogate predicts $P_f$, $E_{avg}$ and $E_{std}$. Parameter selection strategies try different $A$, until it finds a promising $\tilde{A}$. We propose two offline strategies, which do not need to call a QUBO solver. Since now we focus on optimising $A$ for one instance $g$, we drop $g$ and use notation $P_f(A)$, $E_{avg}(A)$ and $E_{std}(A)$ for the sake of simplicity.

\subsubsection{Minimum Fitness Strategy} \label{sec:MFS}

First, we propose Minimum Fitness Strategy (MFS), which attempts to predict the optimal relaxation parameter $\tilde{A}$ that leads to minimum fitness. Given $P_f(A)$, $E_{avg}(A)$ and $E_{std}(A)$, with a Gaussian assumption on the conditional distribution of $E_{avg}(A)$ and $E_{std}(A)$ given $A$, we are able to calculate the expectation of minimum fitness analytically:

\begin{equation}
    \mathcal{E}\left ( \bar{d} \right ) \approx \int_{0}^{\infty } \left (1 - \mathit{\Phi} \left (z; E_{avg}(A), E_{std}(A)^2 \right )  \right )^{P_f(A) \times B} dz
    \label{equ:min_fit}
\end{equation}

where $\mathcal{E}\left ( \bar{d} \right )$ represents the expectation of minimum fitness. $\mathit{\Phi}$ is the Cumulative Distribution Function of Gaussian function. $B$ is the number of soltuions in a batch. (see appendix for the details of derivation and calculation). Therefore the expectation of minimum fitness $D_{min}$ can be seen as a function of $P_f(A)$, $E_{avg}(A)$ and $E_{std}(A)$. Then, the optimal relaxation parameter $\tilde{A}$ can be found by conventional optimisation method.

We use \emph{shgo} optimiser from \emph{scipy} to search parameter search. The search is based on the solver surrogate and does not involve QUBO solver. The search can be done on the CPU within a few seconds.

\subsubsection{$P_f$-based Strategy}

Since $P_f$ provides clue for the location of optimal solution, we can use $P_f$-based Strategy (PBS) to find promising parameters. PBS allows customisable parameter search. The equation for searching the parameters is shown as eq.(\ref{equ:pbs})

\begin{equation}
    \tilde{A}=\underset{A}{\mathrm{argmin}} \; |\mathit{P_f(A)} - p|
    \label{equ:pbs}
\end{equation}

Here, $p$ is a user-defined parameter representing the desired feasibility probability and Eq.(\ref{equ:pbs}) finds parameter $A$ that corresponds to the desired value. As before, it does not involve QUBO solvers. This strategy is useful for a variety of application scenarios. For example, if obtaining a feasible solution in one trial is of primary importance and its objective value is of secondary importance, then $p=90\%$ Would be a reasonable choice. If the user can afford a few trials for each instance, say 5 trials, one could set $p=90\%, 70\%, 50\%, 30\%, 10\%$, and find the corresponding $A$ for each instance.

\section{Travelling Salesman Problem as A Case Study} \label{sec:TSP}

The travelling salesman problem (TSP) is a classical combinatorial optimisation problem, where we are given a list of vertices and their pairwise distances, and we want to visit every vertex exactly once and return to the starting vertex. The goal is to minimize the distance of the tour, i.e., we want to find the shortest Hamiltonian cycle.

\subsection{QUBO Form and Relaxation Parameter}

In \cite{lucas2014ising}, a QUBO formulation that only involves a quadratic number of terms in the number of cities is proposed. Without loss of generality, we can focus on the case where the graph is fully connected, as we can always introduce edges of infinite distances otherwise. We let $d_{uv}$ be the distance between city $u$ and city $v$. We require $n^2$ variables for an $n$-city instance. The first subscript of $x$ represents the city and the second indicates the order that the city is going to be visited at. Let $x_{v,j}$ be the indicator variable that the city $v$ is the $j$-th city to be visited. Notice that the constraint implies that this satisfies the permutation condition. The formulation is as follows:

\begin{equation}
    \min_{x} H_B(x) + AH_A(x)
\end{equation}
where
\begin{equation}
    H_B(x)=\sum_{(u,v) \in E} d_{uv} \sum_{j=1}^n x_{u,j}x_{v,j+1}
\end{equation}
describes the total distance travelled and 
\begin{equation}
    H_A = \sum_{v=1}^n \left( 1-\sum_{j=1}^n x_{v,j}\right)^2 + \sum_{j=0}^n \left( 1-\sum_{v=1}^n x_{v,j}\right)^2
\end{equation}
describes the constraints to be a feasible cycle.

To find the optimal relaxation parameter $A$ for a TSP instance, conventional methods require a few calls to QUBO solvers. Meanwhile QROSS is designed to work without calling a QUBO solver. In order to have comparable experiments, we design an online strategy and evaluate it along with the two offline strategies we proposed in section \nameref{sec:inference}.

\subsection{Online Fitting Strategy for Case Study} \label{sec:ofs}

Online Fitting Strategy (OFS) exploits results of an instance from QUBO solvers to improves parameter search for the same instance. As the relation of $A$ and $P_f$ resembles the sigmoid shape shown in Fig.\ref{fig:pf_dmin}, We can approach promising parameter through the process of curve fitting. The ansatz function to fit is:

\begin{equation}
    S(A, \theta_s, \theta_o ) = \frac{1}{1+e^{-A\theta_s+\theta_o}}
    \label{equ:sigmoid}
\end{equation}

where $\theta_s$ and $\theta_o$ represent the scaling and the offset of the sigmoid shapes in the direction of $A$, respectively. As more hyper-parameters are evaluated, the parameters $\theta_s$ and $\theta_o$ will be estimated more accurately and facilitates the hyper-parameter optimisation better. We use $\mathbb{P}_f(A)$ and $\mathbb{F}(A)$ to represent the ground truth of probability of feasibility and fitness with respect to $A$ found by a QUBO solver. The pseudo algorithm for the Online Fitting Strategy is listed in Algorithm \ref{alg:ofs}.

\begin{algorithm}
\SetAlgoLined
\KwResult{Near optimal parameter $\tilde{A}$}
 Find approximated $A_\textrm{left}$ s.t. $\mathbb{P}_f(A_\textrm{left})=0$\;
 Find approximated $A_\textrm{right}$ s.t. $\mathbb{P}_f(A_\textrm{right})=1$\;
 \While{not reach max num of trials}{
  Fit $S(A, \theta_s, \theta_o )$ with history of $\mathbb{P}_f$\;
  Draw $A_\textrm{next} \sim \mathit{U} \left (A \; | \; 0<S(A, \theta_s, \theta_o )<1\right)$  \;
  Evaluate $\mathbb{P}_f(A_\textrm{next})$ and $\mathbb{F}(A_\textrm{next})$\;
  Update $A_\textrm{left}$ or $A_\textrm{right}$ with $A_\textrm{next}$ if applicable\;
 }
 Return the best $A$ among history of $\mathbb{F}$\;
 \caption{Online Fitting Strategy}
 \label{alg:ofs}
\end{algorithm}

Line 1 and 2 find the left bound and right bound of $\{ A \; | \; 0<\mathbb{P}_f(A)<1 \}$. The bound does not have to be accurate, as a QUBO solver has stochastic nature in its results. The approximated left bound can be found by evaluating a sequence of $\{ A, \frac{A}{2}, \frac{A}{4}, \ldots \}$ on the QUBO solver, until $\mathbb{P}_f(A_\textrm{left})=0$ is found. A similar way applies for the right bound. The offline strategies in section \nameref{sec:inference} is able to provide good guess of initial $A$ without calling QUBO solvers.

The search for $A_\textrm{left}$ and $A_\textrm{right}$ forms a history of $\mathbb{P}_f$, i.e., a collection of points $\{(A_0, \mathbb{P}_f(A_0)), (A_1, \mathbb{P}_f(A_1)), \ldots\}$. The history of $\mathbb{P}_f$ can be used to fit $S(A, \theta_s, \theta_o )$ in Line 4. In Line 5 and 6, A randomly-picked candidate $A_\textrm{next}$ is evaluated on the QUBO solver to find $\mathbb{P}_f(A_\textrm{next})$ and $\mathbb{F}(A_\textrm{next})$. Then $\mathbb{P}_f(A_\textrm{next})$ will join the history of $\mathbb{P}_f$ for the Sigmoid fitting (Line 4) in the next iteration. The loop terminates when maximum iteration is reached.

\section{Experiments} \label{sec:experiments}

In the experiments we compare QROSS with baseline methods on synthetic and real-world datasets. We evaluate the generalisation of QROSS on Fujitsu \emph{DA} and \emph{Qbsolv} from DWave. \footnote{For Qbsolv we are using a simulator backend, instead of a true quantum device backend, due to budget limitation.} All results are summarised in Table \ref{tab:compare}.

\emph{Dataset} We construct a synthetic TSP dataset for the experiments. The whole dataset consists of 300 TSP instances with number of cities ranging from 20-30. The detailed generation method of the dataset is given in the Appendix. We use 270 instances for training the solver surrogate. The remaining 30 instances are used for testing.

\emph{Strategy} One can use any of the strategies we proposed depending on application requirements. For the purpose of benchmarking, we use a mixture of the three strategies in the experiments. The composed strategy is shown as the following list.

\begin{enumerate}
    \item Use MFS to proposes the first candidates
    \item Then use PBS to proposes the next two candidates $\{A \;|\;P_f(A)=80\% \textrm{ and } 20\%\}$
    \item Use OFS to proposes more candidates
\end{enumerate}

The candidates proposed by the composed strategy are evaluated on a QUBO solver in sequence. The $\{A \;|\;P_f(A)=80\% \textrm{ and } 20\%\}$ in the second step is just an example of the usage. One can use other probability setting that suits the application. The trials in the first two step can be used for curve fitting in the third step.

\subsection{Comparison with Generic Methods}

We compare QROSS with three popular methods on the synthetic dataset for testing. The baseline methods are Bayesian Optimisation (BO), Tree-structured Parzen Estimator (TPE), and Random Search. We choose BO and TPE because they are famous for finding optimal solutions with fewer trials. Random Search is also included as a representative of exhaustive methods.

The optimal relaxation parameters of the instances in the synthetic dataset are all within the range $[1, 100]$. Therefore we restrict the exploration of relaxation parameter space to $[1, 100]$ for all baseline methods. BO requires some random samples before the actual exploration and exploitation. For each instance, we draw 5 random samples from a uniform distribution $U(1, 100)$, as we are only comparing the quality of the first 20 trials of baseline methods.

\begin{figure}[htb!]
    \centering
    \includegraphics[width=0.8\columnwidth]{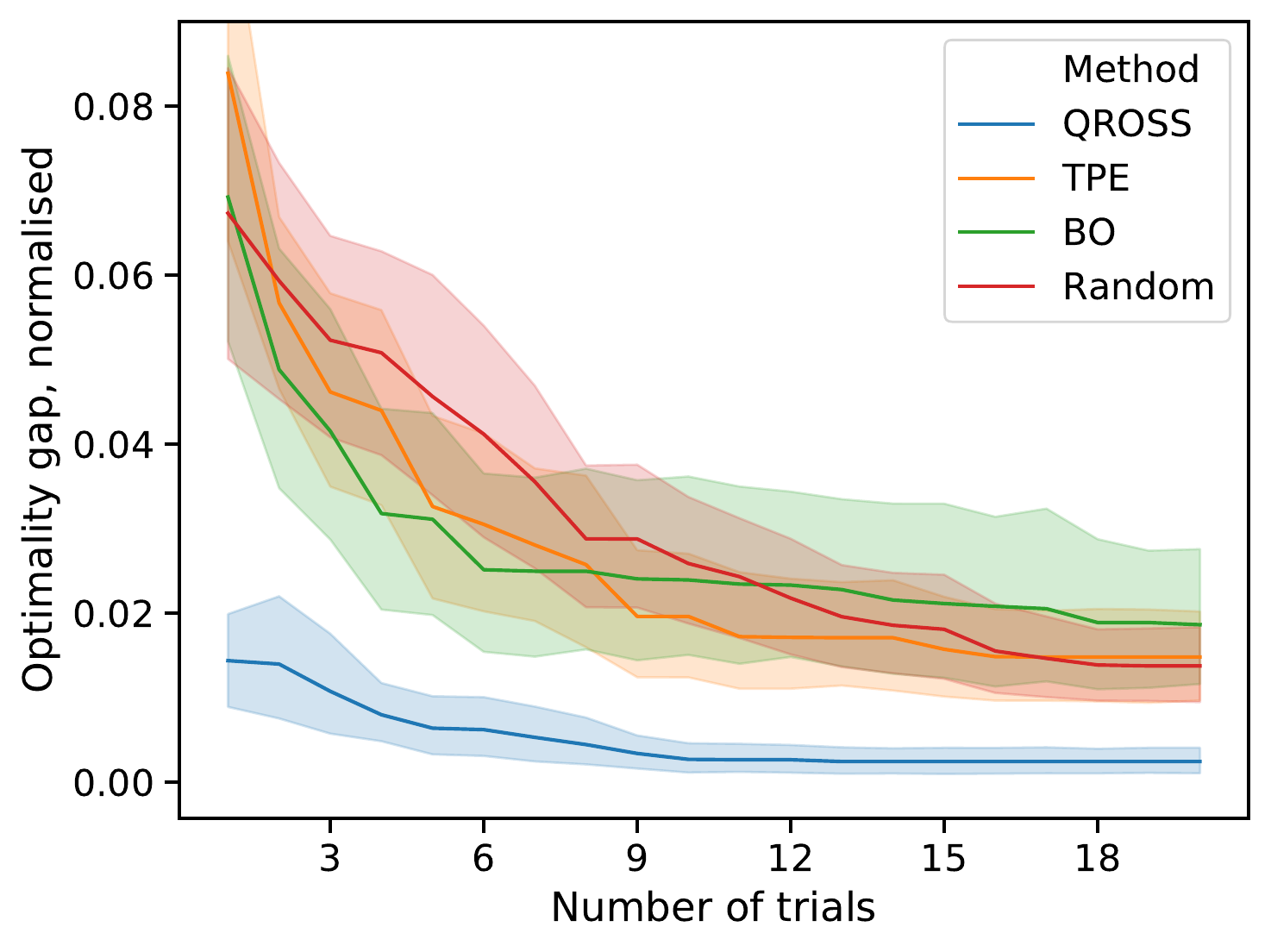}
    \caption{Comparison on test instances of the synthetic dataset. X-axis is the number of trials a method has taken. Y-axis is normalised gap between the near-optimal fitness and the best fitness found so far by a method. The normalised gap is averaged across all test instances.}
    \label{fig:compare_synth}
\end{figure}

Fig.\ref{fig:compare_synth} shows the comparison between QROSS and baseline methods. The blue solid curve for QROSS outperforms other baseline methods by at least $5\%$ at the first trial and $2.9\%$ at the third trial. This means MFS and PBS find better relaxation parameter. We emphasise that the parameters proposed by QROSS so far require no calls to the QUBO solver, whereas TPE and BO improve the parameters based on the results from the QUBO solver. Then the blue curve continue to decline gradually and stay below all other curves. This suggests OFS are also working well on the synthetic dataset. The shade around the curve represents 95\% confidence interval. QROSS has a narrower confidence interval, which suggests that it generalise better to the synthetic dataset than other methods do.

\subsection{Performance on A Real-world Dataset}

\begin{figure}[htb!]
    \centering
    \includegraphics[width=0.8\columnwidth]{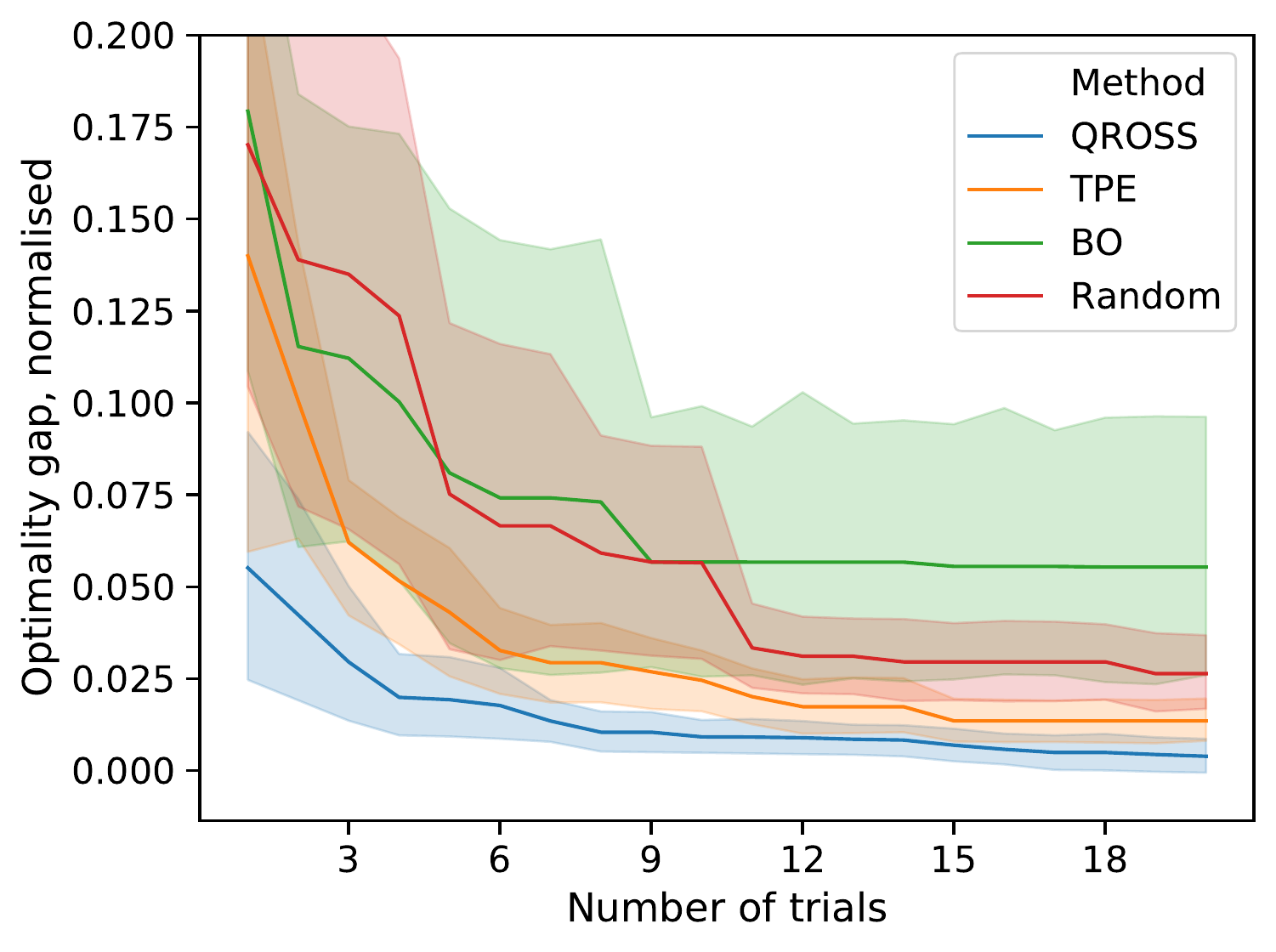}
    \caption{Comparison on TSPLIB dataset}
    \label{fig:compare_tsplib}
\end{figure}

The real-world dataset contains eleven TSP instances from TSPLIB \cite{reinelt1991tsplib}. We exclude TSP instances with number of cities $N>=90$ because decomposition technique is not the focus of this paper. TSP with $N<=14$ is also excluded because they are not challenging enough for QUBO sovlers. We use techniques mentioned in section \nameref{sec:data_prep} to pre-process these instances.  Fig.\ref{fig:compare_tsplib} shows the comparison. The plot settings are the same as Fig.\ref{fig:compare_synth}.

From Fig.\ref{fig:compare_tsplib} we know that the blue curve for QROSS outperforms all other baselines by a large margin of at least $8\%$ at the first trial and 2.9\% at the third trial. We emphasise that the parameters QROSS proposes in the first three trials require no calls to the QUBO solver. for the rest of trials, QROSS with OFS retains its leading position all the way. The 95\% confidence interval of QROSS suggests that its generalisation on the real-world dataset is as good as these baseline methods.

We emphasise that the surrogate is trained on a synthetic dataset, which has problem size of 20-30. The problem size in the real-world dataset is 14-90. QROSS outperforms all baseline methods on this out-of-distribution dataset. This suggests the knowledge in the previous instances generalise well to instances of different size. In other words, QROSS benefits from the learned knowledge about general TSP instances.

\subsection{Generalisation}

\begin{table}[]
\centering
\begin{tabular}{|c|c|c|c|c|c|}
\hline
\multirow{2}{*}{Solver} & \multirow{2}{*}{Method} & \multicolumn{2}{c|}{Synthetic} & \multicolumn{2}{c|}{TSPLIB} \\ \cline{3-6} 
                        &                         & \#3            & \#20           & \#3            & \#20        \\ \hline
\multirow{4}{*}{DA}     & QROSS                   & \textbf{1.7\%} & \textbf{0.2\%} & \textbf{3.3\%} & \textbf{0.2\%}       \\ \cline{2-6} 
                        & TPE                     & 4.6\%          & 1.5\%         & 6.2\%         & 1.4\%       \\ \cline{2-6} 
                        & BO                      & 4.9\%          & 1.9\%         & 11.2\%        & 5.5\%       \\ \cline{2-6} 
                        & Random                  & 5.2\%          & 1.4\%         & 13.5\%        & 2.6\%       \\ \hline
\multirow{4}{*}{Qbsolv} & QROSS                   & \textbf{3.8\%} & \textbf{0.3\%} & \textbf{4.9\%} & \textbf{0.4\%}       \\ \cline{2-6} 
                        & TPE                     & 3.3\%          & 2.3\%         & 5.3\%         & 1.5\%       \\ \cline{2-6} 
                        & BO                      & 5.1\%          & 2.5\%         & 12.7\%        & 6.2\%       \\ \cline{2-6} 
                        & Random                  & 6.8\%          & 1.6\%         & 15.6\%        & 4.4\%       \\ \hline
\end{tabular}
\caption{Comparison of Optimality Gap, Normalised}
\label{tab:compare}
\end{table}

We repeat all previous experiments on Qbsolv. The last four rows of Table \ref{tab:compare} are based on Qbsolv. We construct the training dataset using solutions generated by Qbsolv. The rest of the experiment settings are similar to that for DA. In this experiments QROSS also outperforms all baseline methods on in-distribution and out-of-distribution dataset. This suggest QROSS generalises well to different QUBO solvers.





\section{Conclusion} \label{sec:conclusion}

Relaxation parameter optimisation is important for finding optimal solutions in a QUBO problem. Most existing parameter tuning methods do not exploit knowledge from previous problem instances. In this paper, we frame the parameter optimisation problem as surrogate model minimisation problem, in which the surrogate is learned from instances of a problem. We capture the common structure of the problem and produce optimal relaxation parameter with fewer calls to QUBO solver. 


\bibliographystyle{unsrt}  
\bibliography{reference}

\begin{thebibliography}{10}

\bibitem{kochenberger2014unconstrained}
Gary Kochenberger, Jin-Kao Hao, Fred Glover, Mark Lewis, Zhipeng L{\"u}, Haibo
  Wang, and Yang Wang.
\newblock The unconstrained binary quadratic programming problem: a survey.
\newblock {\em Journal of Combinatorial Optimization}, 28(1):58--81, 2014.

\bibitem{anthony2017quadratic}
Martin Anthony, Endre Boros, Yves Crama, and Aritanan Gruber.
\newblock Quadratic reformulations of nonlinear binary optimization problems.
\newblock {\em Mathematical Programming}, 162(1-2):115--144, 2017.

\bibitem{glover2018tutorial}
Fred Glover, Gary Kochenberger, and Yu~Du.
\newblock A tutorial on formulating and using qubo models.
\newblock {\em arXiv preprint arXiv:1811.11538}, 2018.

\bibitem{brooks2011handbook}
Steve Brooks, Andrew Gelman, Galin Jones, and Xiao-Li Meng.
\newblock {\em Handbook of markov chain monte carlo}.
\newblock CRC press, 2011.

\bibitem{mcgeoch2014adiabatic}
Catherine~C McGeoch.
\newblock Adiabatic quantum computation and quantum annealing: Theory and
  practice.
\newblock {\em Synthesis Lectures on Quantum Computing}, 5(2):1--93, 2014.

\bibitem{aramon2019physics}
Maliheh Aramon, Gili Rosenberg, Elisabetta Valiante, Toshiyuki Miyazawa,
  Hirotaka Tamura, and Helmut~G Katzgraber.
\newblock Physics-inspired optimization for quadratic unconstrained problems
  using a digital annealer.
\newblock {\em Frontiers in Physics}, 7:48, 2019.

\bibitem{goddard2017will}
Phil Goddard, Susan Mniszewski, Florian Neukart, Scott Pakin, and Steve
  Reinhardt.
\newblock How will early quantum computing benefit computational methods?
\newblock In {\em Proc. SIAM Annu. Meeting}, 2017.

\bibitem{sao2019application}
Masataka Sao, Hiroyuki Watanabe, Yuuichi Musha, and Akihiro Utsunomiya.
\newblock Application of digital annealer for faster combinatorial
  optimization.
\newblock {\em Fujitsu Scientific and Technical Journal}, 55(2):45--51, 2019.

\bibitem{hutter2011sequential}
Frank Hutter, Holger~H Hoos, and Kevin Leyton-Brown.
\newblock Sequential model-based optimization for general algorithm
  configuration.
\newblock In {\em International conference on learning and intelligent
  optimization}, pages 507--523. Springer, 2011.

\bibitem{gonzalez2016gpyopt}
J~Gonz{\'a}lez and Z~Dai.
\newblock Gpyopt: a bayesian optimization framework in python, 2016.

\bibitem{snoek2012practical}
Jasper Snoek, Hugo Larochelle, and Ryan~P Adams.
\newblock Practical bayesian optimization of machine learning algorithms.
\newblock {\em Advances in neural information processing systems},
  25:2951--2959, 2012.

\bibitem{bergstra2013hyperopt}
James Bergstra, Dan Yamins, and David~D Cox.
\newblock Hyperopt: A python library for optimizing the hyperparameters of
  machine learning algorithms.
\newblock In {\em Proceedings of the 12th Python in science conference},
  volume~13, page~20. Citeseer, 2013.

\bibitem{akiba2019optuna}
Takuya Akiba, Shotaro Sano, Toshihiko Yanase, Takeru Ohta, and Masanori Koyama.
\newblock Optuna: A next-generation hyperparameter optimization framework.
\newblock In {\em Proceedings of the 25th ACM SIGKDD International Conference
  on Knowledge Discovery \& Data Mining}, pages 2623--2631, 2019.

\bibitem{xu2008satzilla}
Lin Xu, Frank Hutter, Holger~H Hoos, and Kevin Leyton-Brown.
\newblock Satzilla: portfolio-based algorithm selection for sat.
\newblock {\em Journal of artificial intelligence research}, 32:565--606, 2008.

\bibitem{malitsky2014instance}
Yuri Malitsky.
\newblock Instance-specific algorithm configuration.
\newblock In {\em Instance-Specific Algorithm Configuration}, pages 15--24.
  Springer, 2014.

\bibitem{ansotegui2016maxsat}
Carlos Ans{\'o}tegui, Joel Gabas, Yuri Malitsky, and Meinolf Sellmann.
\newblock Maxsat by improved instance-specific algorithm configuration.
\newblock {\em Artificial Intelligence}, 235:26--39, 2016.

\bibitem{cenamor2016ibacop}
Isabel Cenamor, Tom{\'a}s De~La~Rosa, and Fernando Fern{\'a}ndez.
\newblock The ibacop planning system: Instance-based configured portfolios.
\newblock {\em Journal of Artificial Intelligence Research}, 56:657--691, 2016.

\bibitem{hentenryck2009constraint}
Pascal~Van Hentenryck and Laurent Michel.
\newblock {\em Constraint-based local search}.
\newblock The MIT press, 2009.

\bibitem{benoist2011localsolver}
Thierry Benoist, Bertrand Estellon, Fr{\'e}d{\'e}ric Gardi, Romain Megel, and
  Karim Nouioua.
\newblock Localsolver 1. x: a black-box local-search solver for 0-1
  programming.
\newblock {\em 4or}, 9(3):299, 2011.

\bibitem{reinelt1991tsplib}
Gerhard Reinelt.
\newblock Tsplib—a traveling salesman problem library.
\newblock {\em ORSA journal on computing}, 3(4):376--384, 1991.

\bibitem{burkard1997qaplib}
Rainer~E Burkard, Stefan~E Karisch, and Franz Rendl.
\newblock Qaplib--a quadratic assignment problem library.
\newblock {\em Journal of Global optimization}, 10(4):391--403, 1997.

\bibitem{lucas2014ising}
Andrew Lucas.
\newblock Ising formulations of many np problems.
\newblock {\em Frontiers in Physics}, 2:5, 2014.

\bibitem{booth2017partitioning}
Michael Booth, Steven~P Reinhardt, and Aidan Roy.
\newblock Partitioning optimization problems for hybrid classical.
\newblock {\em quantum execution. Technical Report}, pages 01--09, 2017.

\bibitem{vert2019limitations}
Daniel Vert, Renaud Sirdey, and Stephane Louise.
\newblock On the limitations of the chimera graph topology in using analog
  quantum computers.
\newblock In {\em Proceedings of the 16th ACM international conference on
  computing frontiers}, pages 226--229, 2019.

\bibitem{wilkinson1960error}
James~H Wilkinson.
\newblock Error analysis of floating-point computation.
\newblock {\em Numerische Mathematik}, 2(1):319--340, 1960.

\bibitem{gustafson2017end}
John~L Gustafson.
\newblock {\em The End of Error: Unum Computing}.
\newblock CRC Press, 2017.

\bibitem{barends2014superconducting}
Rami Barends, Julian Kelly, Anthony Megrant, Andrzej Veitia, Daniel Sank, Evan
  Jeffrey, Ted~C White, Josh Mutus, Austin~G Fowler, Brooks Campbell, et~al.
\newblock Superconducting quantum circuits at the surface code threshold for
  fault tolerance.
\newblock {\em Nature}, 508(7497):500--503, 2014.

\bibitem{landauer1995quantum}
Rolf Landauer.
\newblock Is quantum mechanics useful?
\newblock {\em Philosophical Transactions of the Royal Society of London.
  Series A: Physical and Engineering Sciences}, 353(1703):367--376, 1995.

\bibitem{pearson2019analog}
Adam Pearson, Anurag Mishra, Itay Hen, and Daniel~A Lidar.
\newblock Analog errors in quantum annealing: doom and hope.
\newblock {\em NPJ Quantum Information}, 5:1--9, 2019.

\bibitem{joshi2019efficient}
Chaitanya~K Joshi, Thomas Laurent, and Xavier Bresson.
\newblock An efficient graph convolutional network technique for the travelling
  salesman problem.
\newblock {\em arXiv preprint arXiv:1906.01227}, 2019.

\bibitem{miki2018applying}
Shoma Miki, Daisuke Yamamoto, and Hiroyuki Ebara.
\newblock Applying deep learning and reinforcement learning to traveling
  salesman problem.
\newblock In {\em 2018 International Conference on Computing, Electronics \&
  Communications Engineering (iCCECE)}, pages 65--70. IEEE, 2018.

\bibitem{khalil2017learning}
Elias Khalil, Hanjun Dai, Yuyu Zhang, Bistra Dilkina, and Le~Song.
\newblock Learning combinatorial optimization algorithms over graphs.
\newblock In {\em Advances in Neural Information Processing Systems}, pages
  6348--6358, 2017.

\bibitem{wu2019learning}
Yaoxin Wu, Wen Song, Zhiguang Cao, Jie Zhang, and Andrew Lim.
\newblock Learning improvement heuristics for solving the travelling salesman
  problem.
\newblock {\em arXiv preprint arXiv:1912.05784}, 2019.

\bibitem{held1970traveling}
Michael Held and Richard~M Karp.
\newblock The traveling-salesman problem and minimum spanning trees.
\newblock {\em Operations Research}, 18(6):1138--1162, 1970.

\bibitem{wang2018distance}
Shengbin Wang, Weizhen Rao, and Yuan Hong.
\newblock A distance matrix based algorithm for solving the traveling salesman
  problem.
\newblock {\em Operational Research}, pages 1--38, 2018.

\end{thebibliography}

\clearpage

\appendix

\section{Ablation Study}

\begin{figure}[htb!]
    \centering
    \includegraphics[width=0.8\columnwidth]{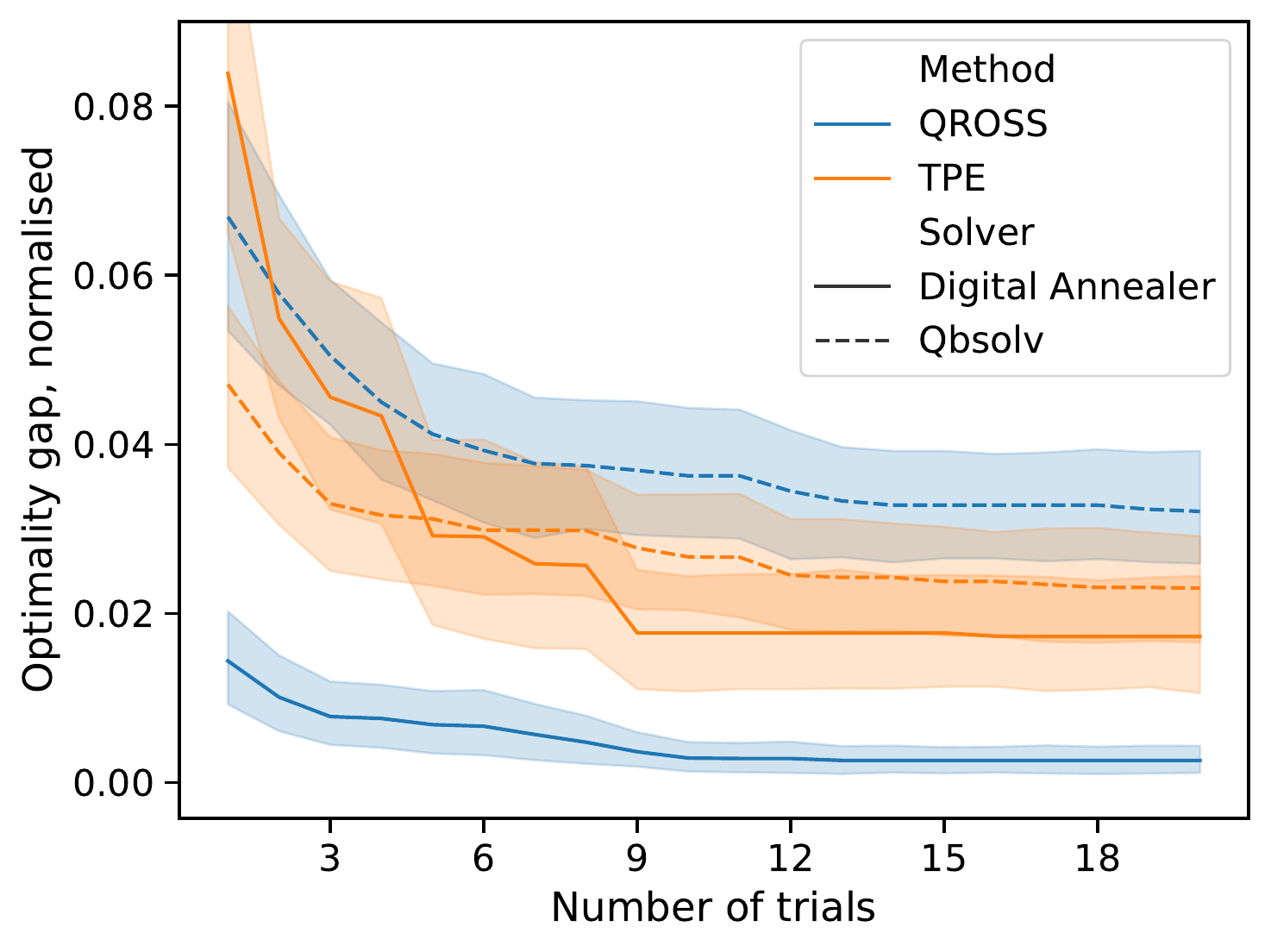}
    \caption{Comparison between QUBO Solvers. All blue curves are for QROSS that trained on DA-based dataset. The blue solid curve is tested with DA, whereas the blue dashed curve is tested with Qbsolv. }
    \label{fig:compare_qubosolver}
\end{figure}

In our experiments, QROSS works well with DA and Qbsolv solver because it is trained on dataset generated by Fujitsu DA \cite{sao2019application} and Qbsolv \cite{booth2017partitioning} from DWave respectively. Next we cross these experiments, i.e., train QROSS on DA-based dataset and test it on Qbsolv, to check if it is something else other than the knowledge in the dataset helps QROSS to outperform all baseline methods. The experiment is carried out on the 30 testing instances of the synthetic dataset. The comparison is shown in Fig.\ref{fig:compare_qubosolver}.

We know from Fig.\ref{fig:compare_qubosolver} that the optimality gap achieved by QROSS on Qbsolv is larger than TPE achieves on Qbsolv. This is because Fujitsu DA and Qbsolv work differently, the knowledge from DA would not be able to generalise to Qbsolv. The performance lag is what we expected for the ablation study.

\section{The choice of Penalty Weight}

Due to the limitation of manufacturing and computation technique, the quantum computers and classical computers we have today are not oracle computers. Errors in these computers could spoils the solutions they produced. In the context of QUBO problem, the choice of penalty weight is critical to the quality of the solutions. We use {\em Minimum Vertex Cover} (MVC) problem to demonstrate the vulnerability of classical computers and quantum computers to large penalty weights.

\subsection{Introduction to {\em Minimum Vertex Cover} (MVC)}

The MVC problem is a classical NP-hard optimisation problem. Given an undirected graph with a vertex set $V$ and an edge set $E$, a vertex cover is a set of vertices such that every edge of the graph has at least one endpoint in this set. A minimum vertex cover is the vertex cover of the smallest size. A standard optimisation model for MVC can be formulated as follows. Let $v_j=1$ if vertex $j$ is in the cover (i.e., in the subset) and $v_j=0$ otherwise. Then the standard optimisation model for this problem is: $$\mathrm{Minimize} \sum_{j\in V} u_j$$ subject to $$u_i+u_j\geq 1, \; \forall (i,j)\in E$$. The constraints ensure that at least one of the endpoints of each edge will be in the cover and the objective function seeks to find the cover using the least number of vertices. 

The constraints in the standard MVC model can be represented by a penalty of the form $\sigma \cdot (1-u_i-u_j+u_iu_j)$. A QUBO form for MVC is $$\mathrm{Minimize} \; \sum_{i\in V}u_i+\sigma\sum_{(i,j)\in E}(1-u_i-u_j+u_iu_j)$$ where $\sigma$ represents a positive scalar penalty. A common extension of this problem allows a weight $w_i$ to be associated with each vertex $i$. The QUBO form for the weighted minimum vertex cover problem is $$\mathrm{Minimize}  \; \sum_{i\in V}w_iu_i+\sigma\sum_{(i,j)\in E}(1-u_i-u_j+u_iu_j)$$ The choice of $\sigma$ is application specific. For example, if $w_i=1, \sigma=2$, adding every additional vertex into a minimum vertex cover will increase the objective energy by three. Removing every vertex from a minimum vertex cover will increase the objective energy by one. Theoretically, any $\sigma > \mathrm{max}(w_i)$ would ensure that a solver can find feasible solutions to the weighted MVC problem. 

\subsection{Quantum and Simulated Annealer on MVC}

\begin{figure}[htb!]
\centering
\includegraphics[width=0.75\linewidth]{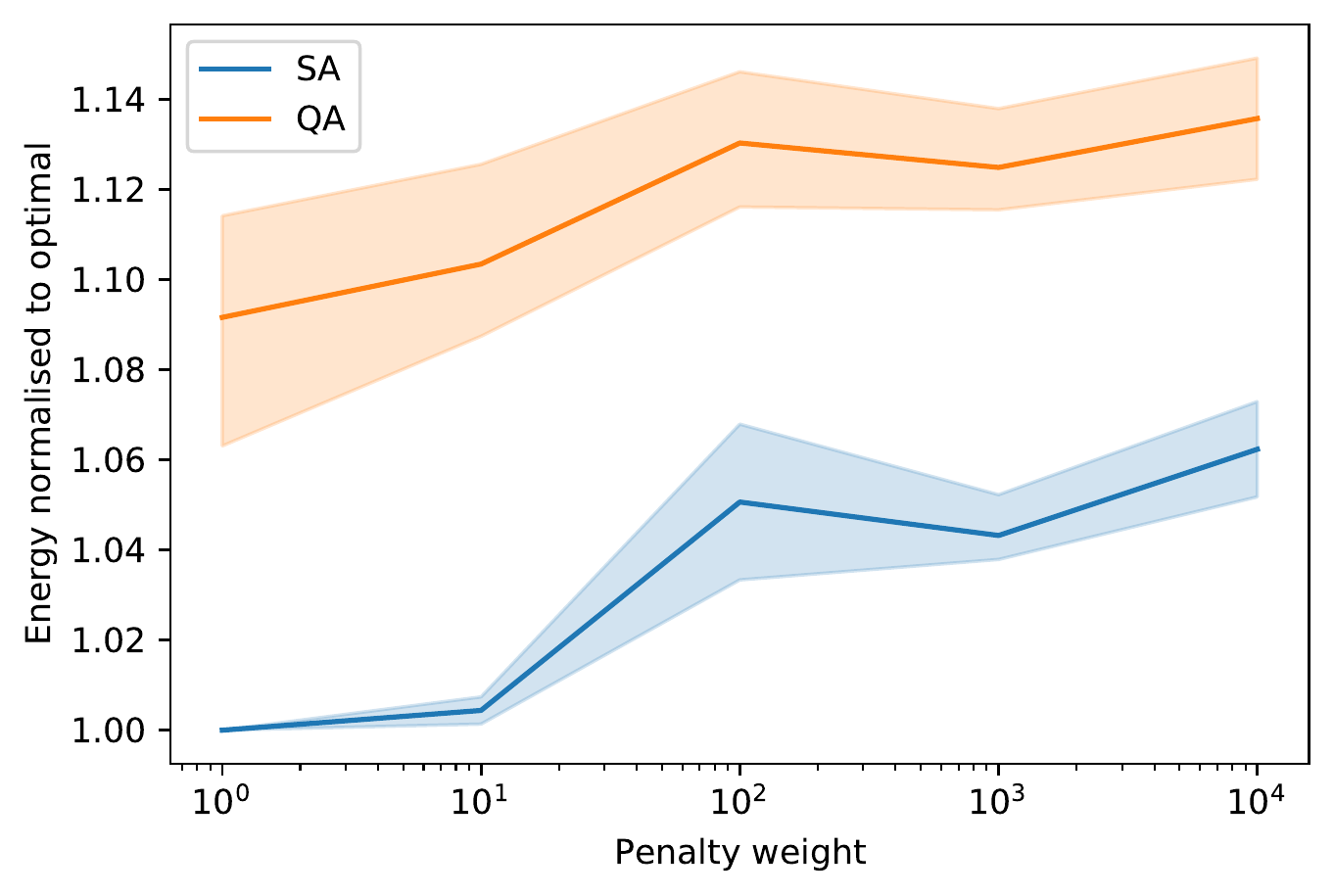}
\caption{Penalty weight v.s. energy normalised. X axis is the penalty weight in log scale. Y axis is the objective energy normalised to the minimum energy state discovered in a run of the experiment. The result is an average over 4 runs for different random seeds. qa is DWave quantum annealer DW\_2000Q. sa is simulated annealing on classical computer. The MVC problem instances are randomly generated graphs with 65 nodes and 50\% probability of connections between any pair of nodes.  The weights of nodes are random numbers following uniform distribution over $[0,1)$.}
\label{fig:analog_control_error}
\end{figure}

We are using Quantum Annealer DW\_2000Q \cite{vert2019limitations} from DWave (qa) and Simulated Annealing on CPU (sa) in this experiment. The MVC problem instances are randomly generated graphs with 65 nodes and 50\% probability of connections between any pair of nodes. \footnote{The number 65 comes from the restrictions of chimera architecture from DWave. The biggest complete graph that can be mapped onto chimera architecture has 65 nodes. In order to ensure a randomly generated graph can be mapped onto the chimera architecture, the number of nodes is set to 65 for our experimental setting.} Fig. \ref{fig:analog_control_error} shows the relation between penalty weight $A$ and normalised objective energy for a MVC problem on DWave DW\_2000Q and Simulated Annealing on a classical computer.

From Fig.\ref{fig:analog_control_error} we understand that the objective energy increases along with the increase of penalty weight. Although this is true for both QA and SA, the reason behind the phenomenon is different for them.

For classical computers, one of the reason is due to the limited range and precision of the numbers and error in floating point operations \cite{wilkinson1960error}. The widely used double precision floating point numbers has \emph{floating point error} in the calculation could accumulates and spoils the results. The floating point error problem is especially prominent when the penalty term dominates the objective energy, as demonstrated in Fig.\ref{fig:analog_control_error}. Fujitsu DA uses FP64. We have similar observation in the main body of the paper as well. The workaround is to tune the penalty weight to mitigate the issue, which one of main focus of this paper. The other is to use Unums representation \cite{gustafson2017end} in the cost of additional hardware/software complexity.

The choice of penalty weight is critical to Quantum computers as well. The imperfections in the quantum hardware is one of the reasons to the degradation in the quality of solutions. More specifically, all physical implementations of quantum computers suffer from analog control errors, in which the coefficients of the Hamiltonian implemented differ from those intended. \cite{barends2014superconducting} If the penalty term dominates the Hamiltonian, the objective of the original problem could be overwhelmed by the analog control error. The analog control error threatens to spoil the results of computations due to the accumulation of small errors. This problem was recognized early on in the gate model \cite{landauer1995quantum} and later in the annealing based model \cite{pearson2019analog}.

\section{Feature Extraction}

In order to learn the common structure of problem instances, we employ deep neural network as it is able to extract feature related to the task in an automatic way. The following literature focus on feature extraction for Traveling Salesman Problem, which is taken as a case study in this paper.

\cite{joshi2019efficient} uses deep Graph Convolutional Networks to build TSP graph representations and output tours in a non-autoregressive manner via beam search algorithm. The representations in \cite{joshi2019efficient} is the probability of each edge being in the route of the optimal solution. \cite{miki2018applying} approaches this edge-level features from a different route. \cite{khalil2017learning} and \cite{wu2019learning} uses reinforcement learning technique to facilitate the heuristic search processing for TSP. 

These methods aim to estimate solutions and/or objective energies, whereas QROSS is for relaxation parameter tuning. We apply the feature extraction technique in \cite{joshi2019efficient,miki2018applying} to capture the common structure shared by TSP.

\section{TSP Problem Generation}

To create augmented data set, We use uniform distribution and exponential distribution as our random number generators to create the coordinates of the cities. The parameter for the exponential distribution is generated from uniform distributions over a range.  The uniform distribution is generated on a bounded domain. After we generated the coordinate data, we then compute the corresponding Euclidean distance.

\section{TSP problem pre-processing and post-processing}

In \cite{held1970traveling}, it is shown that by changing the distance matrix, $$d_{ij}'=d_{ij}-\pi_i - \pi_j,$$, an optimal tour corresponding to the original distance matrix is also the optimal tour for the updated  distance matrix. In \cite{wang2018distance},  Minimizing the Variance of Distance Matrix (MVODM) is proposed where $\pi_i$ is chosen such that the resulting distance matrix has minimal variance. It is shown empirically that such procedure improve greedy search algorithm. We perform such pre-processing to our distance matrix. 


 
After we solve the TSP instance using the solver, we use the original distance matrix to compute the original distance of the original Hamiltonian cycle.

\section{Calculate Expectation of Minimum Fitness}

Suppose the number of solutions in a batch is $B$ for a given problem instance $g$ and a relaxation parameter $A$, the number of feasible solutions in the batch is:

\begin{equation}
    m = P_f(A) \times B
\end{equation}

Let $d_i$ represent the $i$th feasible solution in a batch, $\bar{d}$ is the minimum fitness in batch, then

\begin{equation}
    \bar{d} = \min ( d_0, d_1, \ldots, d_m )
\end{equation}

We define the expectation of minimum distance $\mathit{D_{min}}\left(\cdot\right)$ as a function of $P_f(A)$, $E_{avg}(A)$ and $E_{std}(A)$:

\begin{equation}
    \mathcal{E}\left ( \bar{d} \right ) = \mathit{D_{min}}\left(P_f(A), E_{avg}(A), E_{std}(A)\right)
    \label{equ:dmin_notation}
\end{equation}

where $\mathcal{E}$ stands for expectation operator. All the input of eq.(\ref{equ:dmin_notation}) can be obtained from the output of the surrogate model. Next we are going to find the analytical expression to estimate the expectation of minimum fitness. We use $\Phi \left ( z \right )$ to represent the probability of a fitness $d_i$ in a batch of replicas being smaller than a given value $z$:

\begin{equation}
    \Phi \left ( z \right ) = P\left ( d_i < z \right )
\end{equation}

Then we have the analytical expression of $P\left ( d_i > z \right )$ and $P\left ( \bar{d} < z \right )$:

\begin{equation}
    P\left ( d_i > z \right ) = 1 - \Phi \left ( z \right )
\end{equation}

\begin{equation}
    P\left ( \bar{d} < z \right ) = 1- \left (1 - \Phi \left ( z \right )  \right )^m
\end{equation}

The expectation of $\bar{d}$ can be expressed by:

\begin{equation}
    \mathcal{E}\left ( \bar{d} \right ) = \int z \frac{\partial P\left ( \bar{d}<z \right )}{\partial z}dz
    \label{equ:dmin_exact}
\end{equation}

The fitness $( d_0, d_1, \ldots, d_m )$ are non-negative random variables, which means eq.(\ref{equ:dmin_exact}) could be approximated as:

\begin{equation}
    \mathcal{E}\left ( \bar{d} \right ) \approx \int_{0}^{\infty } 1- P\left ( \bar{d}<z \right ) dz
    \label{equ:dmin_approx}
\end{equation}

The exact distribution of $d_i$ is unknown. We observed in our experiments that the distribution is a bell shape in most of cases. Therefore, we assume the $d_i$ follows Gaussian distribution, which can be expressed in the following equation:

\begin{equation}
    \left ( d_0, d_1, \ldots, d_m \right ) \sim \mathcal{N} \left ( E_{avg}(A), E_{std}(A)^2 \right )
    \label{equ:gaussian}
\end{equation}

Then $\Phi$ could be expressed as the cumulative density function of the Gaussian distribution:

\begin{equation}
    \Phi = \mathit{CDF} \left ( E_{avg}(A), E_{std}(A)^2 \right )
    \label{equ:cdf}
\end{equation}

One can plug eq.(\ref{equ:cdf}) into eq.(\ref{equ:dmin_approx}) to calculate the expectation of minimum fitness. Please note that when $P_f$ is very close to zero, there is no feasible solutions in a batch of replicas. the calculation of $P\left ( \bar{d} < z \right )$ would be meaningless. We set $\lim_{P_f \to 0} \mathit{D_{min}} = +\infty$. Lastly we can solve the following equation using off-the-shelf optimiser to find the optimal hyper-parameter $\tilde{A}$:

\begin{equation}
    \tilde{A}=\underset{A}{\mathrm{argmin}} \; \mathit{D_{min}}(\mathit{P_f(A)},\mathit{E_{avg}(A)},\mathit{E_{std}(A)})
    \label{equ:min_fit_argmin}
\end{equation}

\section{Surrogate model} \label{sec:surrogte}

\begin{figure}[htb!]
    \centering
    \includegraphics[width=0.8\columnwidth]{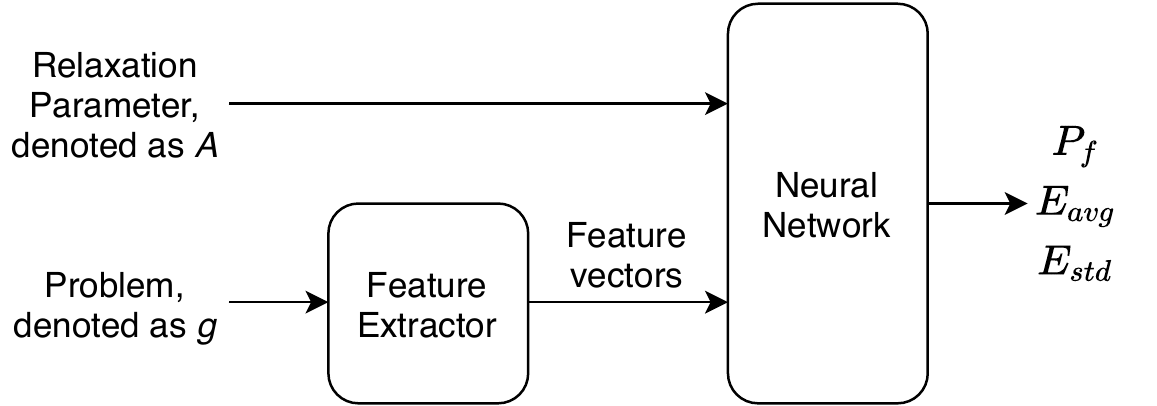}
    \caption{The structure of solver surrogate}
    \label{fig:surrogate}
\end{figure}

The structure of the neural network is shown in Fig.\ref{fig:surrogate}. It takes two inputs: 1. a problem instance $g$; 2. a relaxation parameter $A$. It predict probability of feasibility $P_f$, statistics of objective energy $E_{avg}$ and $E_{avg}$, as functions of $g$ and $A$.

$g$ goes through feature extraction first. The choice of the feature extraction layer depends on the representation of a problem. Possible choices are convolutional networks for matrix representations, graph convolutional networks for graph representations or recurrent networks for sequential representations. Good examples for TSP are \cite{joshi2019efficient,miki2018applying}, in which the authors take adjacency matrix of TSP as fully connected graph and apply graph convolutional neural network to extract features.

Our experiments are based on the pre-trained model for feature extraction in \cite{joshi2019efficient}. The edge-level features cannot be directly applied in our case. Still, we can aggregate the edge-level features into graph-level ones, which is what we needed for QROSS method.

Then the feature vectors of $g$, together with the $A$, goes into fully connected layer for prediction. Since the nature of $P_f$ is different from that of $E_{avg}$ and $E_{std}$, we train these target separately. We use Binary Cross Entropy (BCE) loss for $P_f$ and use Huber loss for $E_{avg}$ and $E_{std}$.

\section{Comparison Between QUBO Solver and Solver Surrogate}

\begin{figure}[htb!]
    \centering
    \includegraphics[width=0.8\columnwidth]{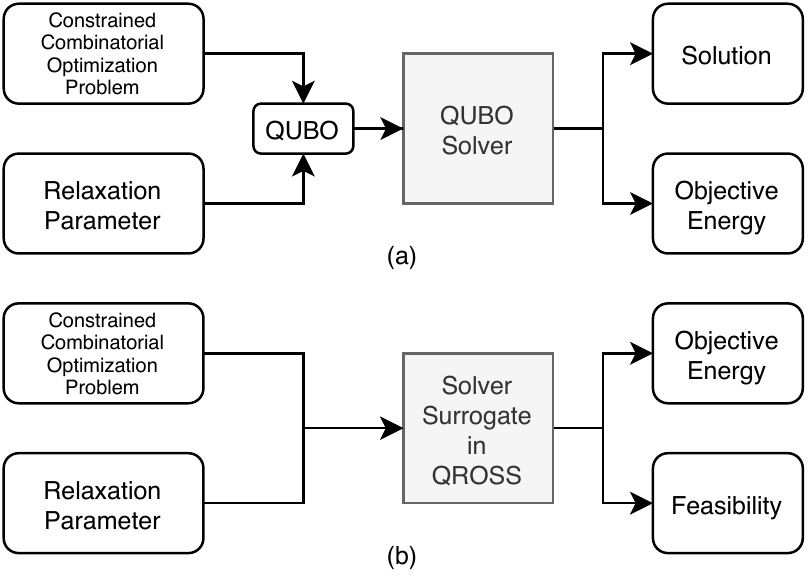}
    \caption{Comparison between a QUBO solver and the solver surrogate in QROSS}
    \label{fig:surrogate_compare}
\end{figure}

Fig.\ref{fig:surrogate_compare} compares a QUBO Solver and the Solver Surrogate in QROSS. A QUBO solver finds exact solutions, given a QUBO problem. The surrogate does not find exact solutions. Instead, it predicts the $P_f$ and objective energy. Because the prediction is much cheaper than the solution-find process in terms of time and energy, QROSS is able to find promising parameters.

\end{document}